\documentclass{svproc}
\usepackage{amssymb,url,amsmath}
\usepackage{lipsum}
\usepackage{graphicx}
\usepackage{hyperref}

\usepackage[ruled,vlined]{algorithm2e}

\newtheorem{conj}[theorem]{Conjecture}

\usepackage{xspace}

\def\MT{\textsc{Min-Time}\xspace}
\def\MD{\textsc{Min-Distance}\xspace}
\def\MI{\textsc{Min-Immersions}\xspace}

\begin{document}
  
\title{Efficient inspection of underground galleries using $k$ robots with limited energy
\thanks{The problem studied in this paper is in the framework of the project ``Algorithms for autonomous navigation of underground systems'' funded by the Company SPT (Stockholm Precision Tools, http://www.stockholmprecisiontools.com/). This research has also received funding from the project GALGO (Spanish Ministry of Economy and Competitiveness, MTM2016-76272-R AEI/FEDER,UE). It has also received funding from the European Union's Horizon 2020 research and innovation programme under the Marie Sk\l{}odowska-Curie grant agreement No 734922. \protect\includegraphics[height=1em]{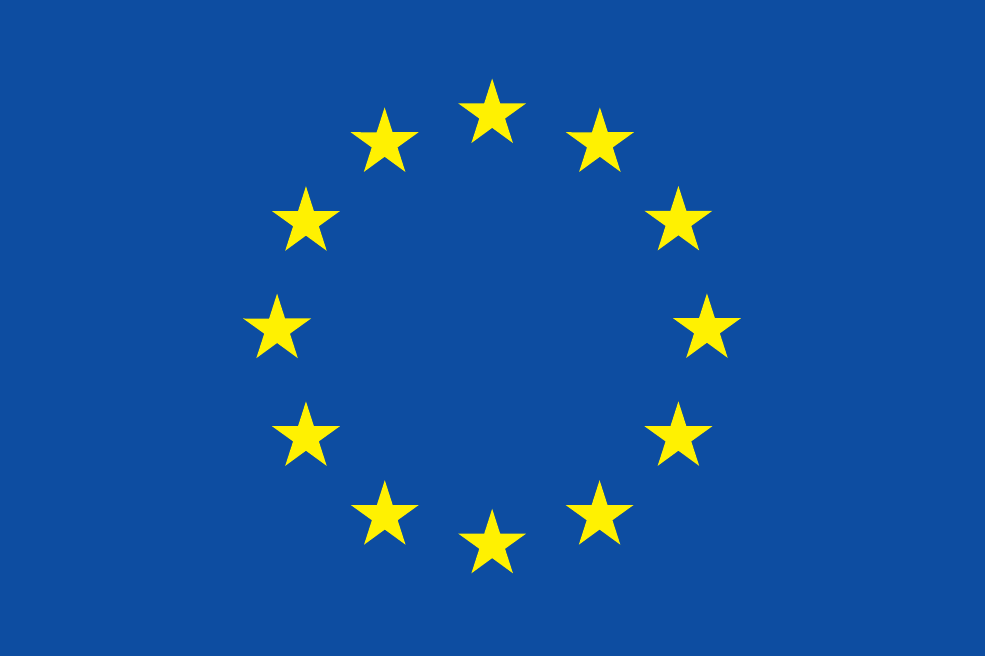}
}
}

\titlerunning{Inspection of underground galleries using $k$ robots with limited energy} 

\author{S. Bereg\inst{1} \and L.E. Caraballo\inst{2} \and J.M. D\'iaz-B\'a\~nez\inst{2}}
%
%

\institute{Univesity of Texas at Dallas, USA\\
\email{besp@utdallas.edu}
\and
University of Seville, SPAIN\\
\email{lcaraballo@us.es, dbanez@us.es}}


\maketitle

\begin{abstract}
We study the problem of optimally inspecting an underground (underwater) gallery 
with $k$ agents. We consider  a gallery with a single opening and with a tree topology rooted at the opening.
Due to the small diameter of the pipes (caves), the agents are small robots with limited autonomy and there is a supply station at the gallery's opening. Therefore,
they are initially placed at the root and periodically need to return to the supply station. Our goal is to design off-line strategies to efficiently cover the tree with $k$ small robots. We consider two objective functions: the covering time (maximum collective time) and the covering distance (total traveled distance). The maximum collective time is the maximum time spent by a robot needs to finish its assigned task (assuming that all the robots start at the same time); the total traveled distance is the sum of the lengths of all the covering walks. 
Since the problems are intractable for big trees, we propose approximation algorithms. Both efficiency and accuracy of the suboptimal solutions are empirically showed for random trees through intensive numerical experiments.
\end{abstract}

%
%

\section{Introduction}

Suppose we want to explore an underground (underwater) gallery by using $k$ small robots with limited power. Then, due to this autonomy constraint, they need to periodically return  outside to recharge battery. We can associate a graph with
the environment, and more concretely a tree rooted at the outside charge station. We assume that all robots know the map road in advance and the goal is to efficiently cover the tree (the tree is covered when all vertices are visited at least once). In this scenario we address several optimization problems that arise when performing an inspection with two criteria: the \emph{total traveled distance} and the \emph{maximum collective time}, i.e., respectively, the sum of the distances traveled by all robots and the maximum time used by any robot from the team.

\begin{figure}[thb]
\centering\includegraphics[scale=0.35]{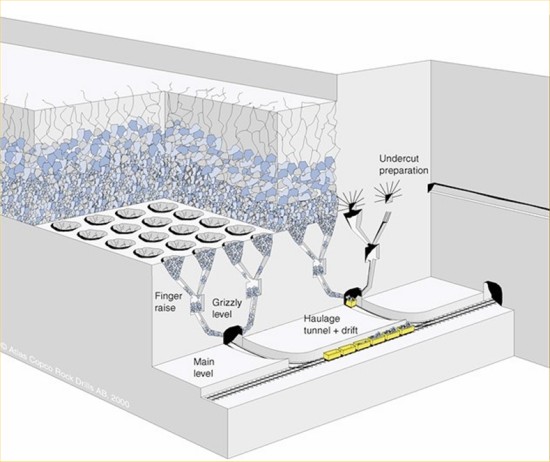}
\caption{An example of a tree in mining \cite{mine}.}
\label{fig:mine}
\end{figure}

The motivation is the inspection of pipelines in mining with robots, the exploration of submarine galleries by using robots or human divers, or rescue tasks in underwater caves when the map road is known in advance. In  Figure~\ref{fig:mine} a gallery with a tree topology is used in the context of the block caving strategy in mining.

Our problems are related to multiple traveling salesman \cite{bektas2006}, or graph exploration problems \cite{arkin2006}. The problem of exploring an environment modeled as a graph has been widely studied, mainly when the environment is unknown (see \cite{rao1993} for a survey). 
In the $k$-Traveling Salesman Problem, all nodes (and edges) of a graph have to be visited (at least once) by one of the $k$ robots initially placed at some node of the graph, subject to minimize the maximum length route of a robot. 
The paper \cite{dynia2006}  shows how to construct (in polynomial time in size of a graph but exponential in $k$) optimal routes for an arbitrary $k$. The main difference in our problems is that the battery power is limited and the robots must return to recharge at the root in at most a fixed amount of time. 

The covering problems using robots with limited energy are NP-hard \cite{ours}, so, there is no efficient known algorithm to solve them. The main contributions of this paper are the study properties of optimal trees and develop efficient sub-optimal algorithms for these problems. Also we develop (non-efficient) optimal algorithms in order to evaluate the results of our sub-optimal algorithms on random trees (we developed an algorithm for generating random trees). Surprisingly, these optimal algorithms have fast running time for trees up to 45 nodes. From this study it turned out that our sub-optimal algorithms perform very well producing covering strategies within a constant approximation.

\section{Problems statement}

Let $T=(V,E)$ be a rooted tree with root $r\in V$.
The \emph{height} of $T$, denoted by $h(T)$, is the maximum distance (number of edges) from $r$ to any node in $T$.
Suppose we have $k$ robots to cover $T$, i.e. to visit all the nodes of $T$ at least once. 
Every agent starts at $r$ and moves from a node to a neighboring one with a unit cost (time or length). Then, we say $T$ is \emph{unweighted}. 
Assume that every agent has a limited autonomy of $p$ moves. 
Also, suppose that the root of $T$ is a \emph{supply station} for the robots, therefore, every time a robot leaving the root can make at most  $p$ moves. 
We assume that $p\ge 2h(T)$, so a robot can visit the farthest node from $r$.  
We call a tour of an agent that starts and ends at $r$, an \emph{immersion}. 
We address the problem of finding (1) a set immersions covering $T$ and (2) an assignment of the immersions the agents such that some cost function is minimized. 
In this work we consider two different cost functions: \emph{maximum collective time}, the maximum time used by any robot from the team, and \emph{cover distance}, total distance traveled by all the agents to cover (inspect) the tree.

Let $\mathcal{I}=\{I_1,\dots,I_m\}$ be a set of $m$ immersions covering the tree $T$. 
Let $S_i\subseteq \mathcal{I}$ be an immersions assignment to an $i$-th agent. 
The cost in time and distance to perform $S_i$ is denoted by $C(S_i)=\sum_{I\in S_i}C(I)$ where $C(I)$ denotes the cost of the immersion $I$ (number of steps in 
the immersion).
We are ready to formally state the problems addressed in this paper.
\begin{problem}\label{pbm:cover-time}
The problem of inspecting an underground tree with $k$ small agents minimizing the maximum collective time:
\begin{flalign*}
&\begin{array}{rl}
\multicolumn{2}{l}{\textsc{Tree-Inspection \MT}}\\
\texttt{Instance:}&\text{A tree }T=(V,E) \text{ rooted at } r\in V,\;k,p\in\mathbb{N} \text{ such that } p\geq 2\cdot h(T)\\
\texttt{Solution:}& \text{A set }\mathcal{I}\text{ of immersions covering }T\; (\forall I\in\mathcal{I}:\;C(I)\leq p)\\
&\text{and a partition }\{S_1,\dots, S_k\}\text{ of }\mathcal{I}\\
\texttt{Goal:}& \text{minimize }\max_{i=1}^k\left\{C(S_i)\right\}
\end{array}&
\end{flalign*}
\end{problem}

\begin{problem}\label{pbm:cover-distance}
The problem of inspecting an underground tree with $k$ small agents minimizing the total traveled distance:
\begin{flalign*}
&\begin{array}{rl}
\multicolumn{2}{l}{\textsc{Tree-Inspection \MD}}\\
\texttt{Instance:}&\text{A tree }T=(V,E),\;r\in V,\;k, p\in\mathbb{N} 
\text{ such that } p\geq 2\cdot h(T)\\
\texttt{Solution:}& \text{A set }\mathcal{I}\text{ of immersions covering }T\; (\forall I\in\mathcal{I}:\;C(I)\leq p)\\
&\text{and a partition }\{S_1,\dots, S_k\}\text{ of }\mathcal{I}\\
\texttt{Goal:}& \text{minimize }\sum_{i=1}^k C(S_i) = \sum_{I\in\mathcal{I}}C(I)
\end{array} & 
\end{flalign*}
\end{problem}

\begin{remark}\label{rmk:distance-allK}
Note that a solution for the goal of Problem~\ref{pbm:cover-distance} is determined by the set of immersions and it does not depend on the immersions assignment. Therefore, given a tree $T$ rooted at a node $r$ and a battery power $p$, the optimal solution of Problem~\ref{pbm:cover-distance} remains invariant for all $k\geq 1$. So we will address this problem with $k=1$.
\end{remark}

\begin{remark}\label{rmk:time-k=1}
It easy to see that problem~\ref{pbm:cover-time} reduces to problem~\ref{pbm:cover-distance} when $k=1$. However, for $k>1$, the problems are essentially different (see Section~\ref{section-properties}).
\end{remark}

Now, we state an auxiliary problem that we will use to solve the problem~\ref{pbm:cover-time} and a new optimization problem related with the classical set cover problem.

\begin{problem}\label{pbm:hybrid}
The problem of inspecting a tree with $k$ agents minimizing the spent time and using a given set of immersions:
\begin{flalign*}
&\begin{array}{rl}
\multicolumn{2}{l}{\textsc{Constrained Tree-Inspection \MT}}\\
\texttt{Instance:}&\text{A tree }T=(V,E),\;r\in V,\;k\in\mathbb{N} \text{\; and\; }p\in\mathbb{N}\text{ such that } p\geq 2\cdot h(T);\\
&\text{A set }\mathcal{I}\text{ of immersions covering }T\; (\forall I\in\mathcal{I}:\;C(I)\leq p).\\
\texttt{Solution:}
&\text{A partition }\{S_1,\dots, S_k\}\text{ of }\mathcal{I}\\
\texttt{Goal:}& \text{minimize }\max_{i=1}^k\left\{C(S_i)\right\}
\end{array} & 
\end{flalign*}
\end{problem}


\begin{problem}\label{pbm:min-immersions} 
Given a tree $T$ and a battery power $p$, compute the minimum number of immersions needed to cover $T$.
\begin{flalign*}
&\begin{array}{rl}
\multicolumn{2}{l}{\textsc{Tree-Inspection \MI}}\\
\texttt{Instance:}&\text{A tree }T=(V,E),\;r\in V,\;k\in\mathbb{N} \text{\; and\; }p\in\mathbb{N}\text{ such that } p\geq 2\cdot h(T)\\
\texttt{Solution:}& \text{A set }\mathcal{I}\text{ of immersions covering }T\; (\forall I\in\mathcal{I}:\;C(I)\leq p)\\
&\text{and a partition }\{S_1,\dots, S_k\}\text{ of }\mathcal{I}\\
\texttt{Goal:}& \text{minimize }|\mathcal{I}|
\end{array} & 
\end{flalign*}
\end{problem}

\section{Properties of optimal solutions}\label{section-properties}

\begin{figure}
\centering
\begin{tabular}{c c c}
	\includegraphics[scale=0.6]{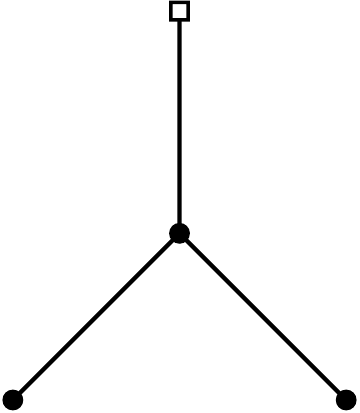}
	&
	\includegraphics[scale=.6, page =2]{tree-exploration.pdf}
	&
	\includegraphics[scale=.6, page =3]{tree-exploration.pdf}
	\\
    (a) & (b) & (c)\\
\end{tabular}
\caption{(a) Tree sample. (b) Immersion of a single agent to cover the tree. (c) The immersions of two agents to cover the tree.}
\label{fig:tree_sample}
\end{figure}

In this section we introduce some notation and properties that will be useful in the design of our algorithms.
First of all, it is easy to see that the problems~\ref{pbm:cover-time} and ~\ref{pbm:cover-distance} are essentially different. 
Consider the tree of Figure~\ref{fig:tree_sample}a and suppose that we have a team of two agents with autonomy 6 or greater. The solution of Problem~\ref{pbm:cover-distance} is shown in Figure~\ref{fig:tree_sample}b where we use just one agent covering the tree with an immersion of length 6 in 6 time units. By other hand, the solution of Problem~\ref{pbm:cover-time} is shown in Figure~\ref{fig:tree_sample}c, where the two agents of the team are used instead. Each agent performs an immersion of length 4, so the total traveled distance is 8 and the maximum collective time is 4.

We need some notation.
In a tree $T=(V,E)$ rooted at $r$, every immersion $I$ determines a subtree $(V_I,E_I)$ rooted at $r$ where $V_I\subseteq V$ is the set of visited nodes in the immersion $I$ and $E_I\subseteq E$ is the set of traversed edges. Note that $2|E_I|=C(I)$. In the following we denote an immersion $I$ by the corresponding subtree, that is, $I=(V_I,E_I)$.
The \emph{leaves nodes} of a tree are the nodes implied in only one edge. In the rest of this work, by convenience, we will only consider the leaves in $V\setminus\{r\}$. We call \emph{inner node} to every non-leaf node in $V\setminus\{r\}$.

The following properties hold for the stated problems.

\begin{lemma}\label{lem:leaves}
The leaves of each immersion $I$ of an optimal solution for  problem \MD  are leaves in the original tree $T$. 
For problems \MT and \MI, there is always an optimal solution with the same property. 
\end{lemma}

\begin{proof}
Suppose that an immersion $I$ (for one of the problems) has a leaf $v$ which is not a leaf of $T$. 
Let $v'$ be a leaf of $T$ such that $v$ is in the path from $r$ to $v'$ and let $I'$ be an immersion containing $v'$. 
Then $I'$ contains $v$ and we can remove $v$ from $I$. 
If the problem is \MD, it is a contradiction. If the problem is \MT or \MI, we repeat this argument to find a solution with the desired property.
\end{proof}

\begin{lemma}\label{lem:leaves2}
Every leaf of $T$ is in exactly one immersion of an optimal solution for problem \MD. 
For problems \MT and \MI, there is always an optimal solution with the same property. 
\end{lemma}

\begin{proof}
Suppose that two immersions $I$ and $I'$ (for one of the problems) have a common leaf $v$ and an edge $\{w,v\}$. We can remove node $v$ and edge $\{w,v\}$ from $I'$ (keeping the same $I$). If the problem is \MD, it is a contradiction. If the problem is \MT or \MI, we repeat this argument to find a solution with the desired property.
\end{proof}

The following corollary is deduced from Lemmas \ref{lem:leaves} and \ref{lem:leaves2}:

\begin{corollary}\label{cor:leaves-immersion}
Any problem \MD, \MT or \MI has a solution such that the leaves of the immersions form a partition of the set of leaves of $T$.
\end{corollary}

\begin{figure}
\centering\includegraphics[scale=0.9]{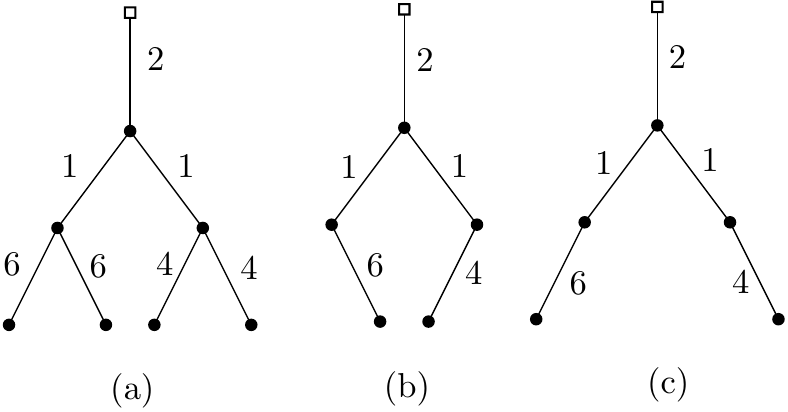}
\caption{(a) A tree $T$. (b) and (c) show two trees in the optimal solution of the  \textsc{Min-Distance} problem for $T$.}
\label{fig:mindist}
\end{figure}

In the following it is showed that we must be careful in designing algorithms for the proposed optimization problems. Many natural properties fail. 
For example, one can approach the \MD problem by covering the leaves of $T$ by disjoint subtrees $T_i$ of $T$ such that
the immersions will be formed by these subtrees and the paths between the roots of $T_i$ and $T$.
Unfortunately, this approach does not work. For example, the tree shown in Figure~\ref{fig:mindist}a has only one optimal solution (non-disjoint) for $p=28$, shown in  Figure~\ref{fig:mindist}b and \ref{fig:mindist}c.

Even more, one can expect that all Min-Distance solutions for the same tree must have the same number of immersions. To see a counterexample of this property, we use the tree from Figure~\ref{fig:mindist}a with a modification that the length of the top edge is 1 and take $p=26$. The same 2 immersions shown in Figure~\ref{fig:mindist}b and \ref{fig:mindist}c make an optimal solution with total cost $26+26=52$. But there exist 3 immersions with the same total cost: two trees of cost $2(1+1+6)=16$ each and one tree of cost $2(1+1+4+4)=20$.

\section{Algorithms}
\label{s:alg1}

The stated optimization problems are all  NP-hard, even the problem of minimizing the maximal distance traveled by a single robot in an unweighted tree\cite{ours}.
In this section we propose optimal and suboptimal algorithms based on the above properties that enable to encode every feasible solution using a partition of the set of leaves of $T$. Let $L=\{l_1,l_2,\dots \}$ be the set of leaves in the given tree $T$. Ours algorithms construct a solution from partitioning the set $L$ (Corollary~\ref{cor:leaves-immersion}). The immersion corresponding to the subset $\{l_{i_1},l_{i_2},\dots\}$ of $L$ is the tree rooted at $r$ formed by the nodes and edges in the paths from $r$ to each one of the leaves $l_{i_1}$, $l_{i_2}$, \dots (Lemma~\ref{lem:leaves}). For convenience, in the rest of this section we denote an immersion $I$ by its corresponding set of leaves.

\subsection{Suboptimal algorithms for the total traveled distance}

For this problem we have a studied and compared two heuristic algorithms. Also, we compare their performance with the optimal solution. By Remark~\ref{rmk:distance-allK} we focus on finding the optimal set of immersions for a single robot. 

\begin{algorithm}
	\SetKwInOut{Input}{Input}
    \SetKwInOut{Output}{Output}    
    \Input{tree $T=(V,E)$ rooted at $r$, autonomy value $p$}
    \Output{set of immersions $\mathcal{I}$}
    $L\gets$ leaves of $T$ in deep-first traversal order\;
    $\mathcal{I}\gets\{\}$;
    $I\gets\{\}$\;
    \ForEach{$l$ in $L$}{
    	\uIf {immersion $I\cup\{l\}$ is feasible with autonomy $p$}{
        	$I\gets I\cup\{l\}$\;
        }
        \Else{
        	$\mathcal{I}\gets \mathcal{I}\cup\{I\}$;
            $I\gets\{l\}$\;
        }
    }
    \KwRet $\mathcal{I}\cup\{I\}$\;
    \caption{Heuristic algorithm \texttt{sweepingLeaves}.}
    \label{alg:sweeping}
\end{algorithm}

Our first heuristic, \texttt{sweepingLeaves}, is based in the following ideas: Let $L=(l_1,l_2,\dots,l_s)$ be a labeling of the leaves obtained from a deep-first traversal of $T$ (starting at the root $r$). Let $I=\{l_i,l_{i+1},\dots,l_{i+c}\}$ be an immersion determined by a subsequence of consecutive leaves in $L$; the cost of $I$ is $C(I) = d(r,l_i)+d(l_i,l_{i+1})+\dots+d(l_{i+c-1},l_{i+c})+d(l_{i+c},r)$ where $d(v,w)$ denotes the distance (number of edges) between the nodes $v$ and $w$. We say that $I$ is feasible if $C(I)\leq p$ and it is not, otherwise.

The goal of \texttt{sweepingLeaves} is to construct a list of immersions of the form: $(I_1=\{l_1,l_2,\dots,l_{c}\},\;I_2=\{l_{c+1},l_{c+2},\dots\,\},\;\cdots\;,\;I_m=\{l_{i},\dots,l_{s-1},l_s\})$
where the leaves of immersion $I_j$ are consecutive to the leaves of $I_{j-1}$ and every immersion is maximal, i.e., if $I_j=\{l_i,l_{i+1},\dots,l_{i+c}\}$ and $l_{i+c}\neq l_s$ then the immersion with leaves $\{l_i,l_{i+1},\dots,l_{i+c},l_{i+c+1}\}$ is not feasible due to the energy constraint. Algorithm~\ref{alg:sweeping} shows a pseudo-code of this heuristic. 


\begin{algorithm}
	\SetKwInOut{Input}{Input}
    \SetKwInOut{Output}{Output}    
    \Input{tree $T=(V,E)$ rooted at $r$, autonomy value $p$}
    \Output{set of immersions $\mathcal{I}$}
    $N_L\gets$ set of leaves of $T$ \;
    $\mathcal{I}\gets\{\}$;
    $I\gets\{\}$\;
    \While{$N_L$ is not empty}{
    	$d_l\gets$ deepest leaf in $N_L$\;
        $I\gets\{d_l\}$;
        $N_L\gets N_L\setminus\{d_l\}$\;
        
    	\While{$N_L$ is not empty}{
        	$l\gets$nearest leaf in $N_L$ to the tree determined by $I$\;
            \uIf{immersion $I\cup\{l\}$ is feasible with autonomy $p$}{
            	$I\gets I\cup\{l\}$;
                $N_L\gets N_L\setminus \{l\}$\;
            }
            \Else{
            	break\;
            }
        }
        $\mathcal{I}\gets\mathcal{I}\cup\{I\}$\;
    }
    \KwRet $\mathcal{I}$\;
    \caption{\texttt{DFTN (deepest-first-then-nearest)}.}
    \label{alg:moya-heuristic}
\end{algorithm}

The second heuristic, which we call \texttt{DFTN (deepest-first-then-nearest)}, is based in the following approach: 
Suppose that we have $j-1$ computed immersions 
 and let $N_L$ be the set of the leaves of the tree that are no involved in any of the computed immersions. If $N_L$ is not empty, starts the $j$-th immersion, denoted by $I_j$, with the deepest leaf in $N_L$ and remove this leaf from $N_L$. Then, if $N_L$ is not empty, try to increase $I_j$ by adding the nearest leaf $l$ of $N_L$ to the tree determined by $I_{j}$. If the immersion $I_{j}\cup\{l\}$ is feasible add the leaf $l$ to $I_j$ and remove it from $N_L$ and repeat the process until $N_L$ is empty or there is no way to increment $I_j$. After that, repeat the whole process to compute the ${j+1}$-th immersion and so on. Algorithm~\ref{alg:moya-heuristic} shows a pseudo-code of this heuristic.

We can make implementations that take linear time for Algorithm~\ref{alg:sweeping} and quadratic time for Algorithm~\ref{alg:moya-heuristic}. We do not include in this work the details of these implementations and the study on their time complexities due to the space restriction.

\subsection{Optimal algorithms for \MD and \MI}
Suppose that we want to compute an exact solution of the \MD problem. One can think of using \texttt{sweepingLeaves} for all possible permutations of the leaves of $T$. This algorithms may fail to find an exact solution (see Figure~\ref{fig:mindist} for example). A correct way would be to test all partitions of $|L|$ leaves. The  running time of this algorithm is $O(B_{|L|}\cdot n)$ where $B_m$ is the $m$-th Bell number \footnote{In combinatorial mathematics, the Bell numbers count the number of partitions of a set.}. The Bell numbers   grow very fast, for example $B_{19}=5,832,742,205,057$.
Thus, it can only work for $n<19$. 
For larger values of $n$, one needs a different approach.
In this Section we propose a faster algorithm using a branch-and-cut paradigm. The pseudocode is shown in Algorithm~\ref{alg:BC}. 
We can apply the same approach to solve \textsc{Min-Immersions}. The only difference is that we store the number of immersions instead of the distance. 

\begin{algorithm}[h]
	\SetKwInOut{Input}{Input}
    \SetKwInOut{Output}{Output}  
    \SetKwFunction{algo}{main}
    \SetKwProg{fmain}{Function}{}{} 
    \SetKwFunction{newtree}{newTree}
    \SetKwFunction{newnode}{newNode}
    \SetKwProg{subroutine}{Procedure}{}{} 
    \Input{tree $T=(V,E)$ rooted at $r$, autonomy value $p$}
    \Output{set of immersions $\mathcal{I}$}
    \fmain{\algo{T,p}}{
    $L\gets$ leaves of $T$ sorted by depth in decreasing order\;
    $\mathcal{I}\gets\{\}$;\qquad $best\mathcal{I}\gets\{\}$\;
    \ForEach{$l$ in $L$}{
    $l$.covered$\gets$false\;
    }
    $optD\gets\infty$; // optimal distance\\
    \newtree{L}\; 
    \KwRet $best\mathcal{I}$\;
    }{}
    \setcounter{AlgoLine}{0}  	
    
    \subroutine{\newtree{L}}{
      Let $l$ be the first uncovered leaf in $L$\;
      \uIf{$l$ is null} 
      {
      $D \gets$ the cost of current immersions in $\cal{I}$\;
      \uIf{D $<$ optD}{
      	$optD\gets D$;\qquad $best\mathcal{I}\gets\mathcal{I}$\;
      }      
      \KwRet\;}
      $I\gets\{l\}$;\qquad // new immersion\\ 	
      \newnode{L,I,l.next}\;
    }{}
    
    \setcounter{AlgoLine}{0}  
    \subroutine{\newnode{L,I,l}}{
    	\uIf {$l$ is null}{
        \uIf{optD $>$ the cost of current immersions in $\cal{I}$}{
        $\mathcal{I}\gets\mathcal{I}\cup\{I\}$\;
        \newtree{L}\;
        $\mathcal{I}\gets\mathcal{I}\setminus\{I\}$\;}
        \KwRet\;}
        \uIf{$C(I\cup \{l\})\le 2p$}{
        $l$.covered$\gets$true;\qquad $I\gets I\cup \{l\}$\; 
        \newnode{L,I,l.next}\; 
        $l$.covered$\gets$false;\qquad $I\gets I\setminus\{l\}$\;
        } 
        \newnode{L,I,l.next}\;
    }{}
  
    \caption{Algorithm \texttt{B-and-C}.}
    \label{alg:BC}
\end{algorithm}

\subsection{Minimizing the maximum collective time}
\label{sec:min_time}
 The heuristic proposed for this problem is a combination of heuristics for problems~\ref{pbm:cover-distance} and \ref{pbm:hybrid}. We generate a sub-optimal solution with a heuristic for  \ref{pbm:cover-distance} and set it as the input for problem~\ref{pbm:hybrid}. The output will be a sub-optimal solution for \ref{pbm:cover-time}.

Problem~\ref{pbm:hybrid} is exactly the so-called multiprocessor scheduling problem for identical processors, one of the most challenging
problems in parallel computing. Given a number of tasks, their execution times and a number of processors,
the goal of multiprocessor scheduling is to
find an assignment to minimize the overall
execution time. In
multiprocessor scheduling problem, a given
program is to be scheduled in a given
multiprocessor system such that the program’s
execution time is minimized, that is, the last job must be
completed as early as possible. 
This problem is intractable
and many heuristics have been proposed to find sub-optimal solutions.
See \cite{khan1994comparison} for a comparison of heuristics and \cite{davis2011survey} for a comprehensive survey on this topic. We use a pseudo-polynomial dynamic programming algorithm for computing the optimal partition of the set of immersions given by the heuristic DFTN. We omit the pseudocode in this version due to the space constraint.




\section{Computational results}
\label{s:results}

All the implementations were performed in MATLAB R2016a (9.0.0.341360) on Linux Mint Cinnamon 64-bit with 16Gb of RAM and a processor Core i7-4720HQ.
We implemented our algorithms from Section 4 and run them on random trees generated as follows. 
%
Suppose that we want to generate a tree of $n$ nodes with labels $\{1,2,\dots,n\}$. We start with the sets, $V=\{1\}$, $S=\{2,\dots,n\}$ and $E=\{\}$ (empty set). Then we apply the following: randomly select an element $v\in V$ and another $w\in S$, then add $\{v,w\}$ to $E$, remove $w$ from $S$ and add it to $V$. Repeat this process until $S$ is empty. The resulting graph is a tree and we select node 1 as the root.

\begin{figure}[h]
\centering\includegraphics[width=\textwidth]{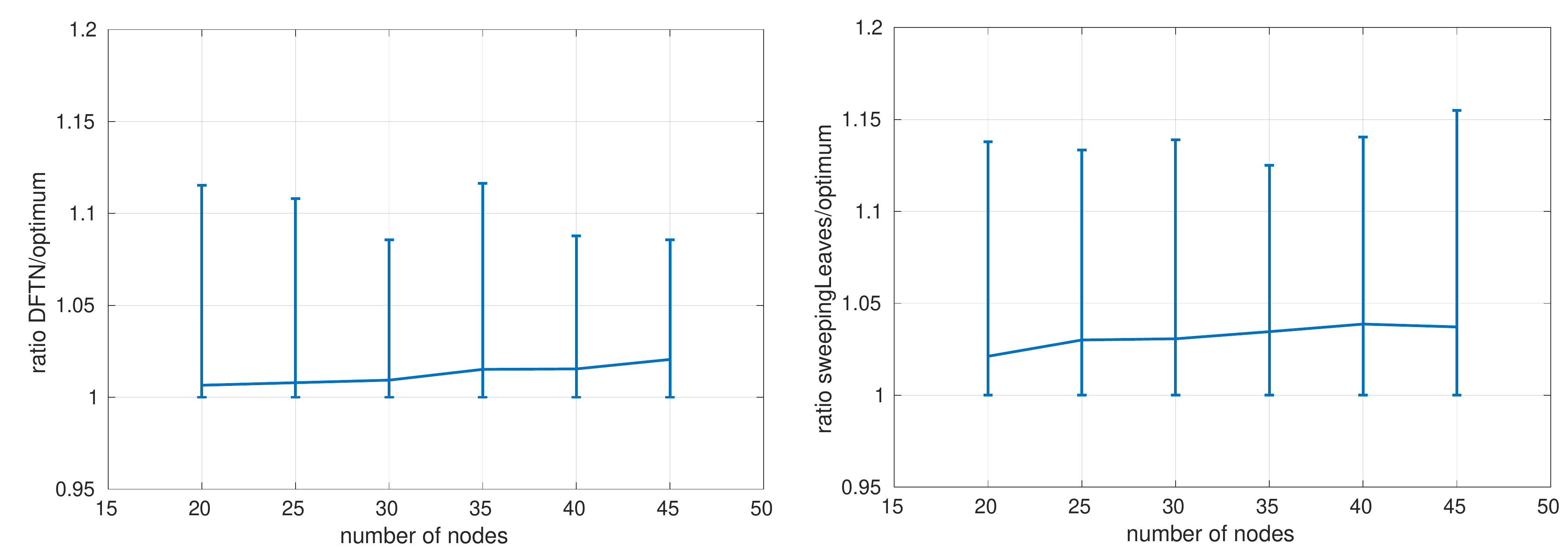}
\caption{Approximation ratio of \texttt{deepest-first-then-nearest} (left) and \texttt{sweepingLeaves} (right) respect to covering distance.}
\label{intervals}
\end{figure}

The computational results are shown in Table 1. 
DFTN heuristic shows better performance in most cases for both \MD
and \MI. 
Figure \ref{intervals} depicts approximation ratios of these algorithms for $n=20,25,..,45$. The vertical interval for each $n$ shows the best and the worst ratios over 100 random trees. The plotted function corresponds to the average ratio. Both algorithms have approximation ratio at most 1.2 and the average ratio less than 1.05. 

\begin{table}[h]
\centering
\begin{tabular}{|cc|cc|cc|cc|cc|c||cc|cc|cc|cc|c|} 
\cline{3-20}
\multicolumn{2}{c|}{} & \multicolumn{9}{c||}{autonomy value $p=2*h$ } & \multicolumn{9}{c|}{autonomy value $p=2*(h+1)$ } \\
\cline{3-20}
\multicolumn{2}{c|}{} & \multicolumn{2}{c|}{\texttt{B\&C} MD} & \multicolumn{2}{c|}{\texttt{B\&C} MI} & \multicolumn{2}{c|}{DFTN}& \multicolumn{2}{c|}{Swp.L} & &   \multicolumn{2}{c|}{\texttt{B\&C} MD} & \multicolumn{2}{c|}{\texttt{B\&C} MI} & \multicolumn{2}{c|}{DFTN}& \multicolumn{2}{c|}{Swp.L}& \\
\hline
$l$ & 
$h$ & Dist & Im & Dist & Im & Dist & Im & Dist & Im & mT & Dist & Im & Dist & Im & Dist & Im & Dist & Im & mT
\\ \hline
17 & 6 & 84 & 9 & 84 & 9 & 88 & 10 & 94 & 11 & 42  & 74 & 8 & 78 & 8 & 78 & 8 & 78 & 8 & 38\\
13 & 7 & 80 & 6 & 82 & 6 & 80 & 6 & 82 & 7 & 42 & 76 & 5 & 78 & 5 & 76 & 5 & 80 & 6 & 44\\
14 & 5 & 86 & 11 & 86 & 11 & 86 & 11 & 88 & 12 & 44  & 74 & 9 & 74 & 9 & 76 & 9 & 76 & 10 & 38\\
15 & 5 & 66 & 8 & 66 & 8 & 66 & 8 & 66 & 8 & 34  & 64 & 7 & 64 & 7 & 64 & 7 & 64 & 7 & 32\\
16 & 6 & 80 & 8 & 80 & 8 & 80 & 8 & 80 & 8 & 40  & 70 & 6 & 76 & 6 & 72 & 6 & 72 & 6 & 36\\
16 & 7 & 76 & 6 & 82 & 6 & 76 & 6 & 82 & 7 & 42  & 74 & 5 & 74 & 5 & 74 & 5 & 76 & 5 & 42\\
14 & 6 & 70 & 9 & 70 & 9 & 74 & 10 & 70 & 9 & 36  & 66 & 9 & 68 & 9 & 70 & 9 & 66 & 9 & 34\\
18 & 7 & 88 & 11 & 90 & 11 & 90 & 11 & 94 & 11 & 44  & 74 & 9 & 74 & 9 & 74 & 9 & 76 & 9 & 38\\
18 & 6 & 128 & 11 & 128 & 11 & 128 & 11 & 138 & 13 & 68  & 96 & 8 & 98 & 8 & 96 & 8 & 102 & 8 & 48\\
12 & 9 & 66 & 4 & 70 & 4 & 68 & 4 & 72 & 5 & 36  & 66 & 4 & 68 & 4 & 68 & 4 & 66 & 4 & 34\\
16 & 6 & 72 & 7 & 72 & 7 & 80 & 8 & 76 & 8 & 36  & 68 & 6 & 68 & 6 & 74 & 7 & 72 & 7 & 38\\
14 & 6 & 78 & 9 & 80 & 9 & 80 & 9 & 80 & 9 & 40  & 70 & 7 & 70 & 7 & 70 & 7 & 72 & 8 & 38\\
14 & 6 & 64 & 7 & 64 & 7 & 64 & 7 & 68 & 8 & 32  & 64 & 7 & 68 & 7 & 66 & 7 & 66 & 7 & 32\\
11 & 5 & 70 & 8 & 70 & 8 & 70 & 8 & 70 & 8 & 36  & 66 & 6 & 66 & 6 & 68 & 7 & 68 & 7 & 34\\ 
17 & 4 & 70 & 11 & 70 & 11 & 70 & 11 & 74 & 12 & 36  & 68 & 10 & 70 & 10 & 70 & 10 & 70 & 10 & 34\\
\hline
\end{tabular}
\vspace{1mm}
\caption{Table of results of our algorithms on random trees with 30 nodes. One row corresponds to a single tree; $l$: number of leaves, $h$: height of the tree,
\texttt{B\&C} MD: optimal algorithm (branch and cut) for \MD,
\texttt{B\&C} MI: optimal algorithm (branch and cut) for \MI,
DFTN: \texttt{deep-first-then-nearest} heuristic and
Swp.L: \texttt{sweepingLeaves} heuristic.
For each one of these algorithms, the columns Dist and Im show the traveled distance and number of immersions, respectively. The column mT shows the suboptimal value of Problem~\ref{pbm:cover-time} for $k=2$ robots by using the heuristic proposed in subsection~\ref{sec:min_time}.}
\end{table}

\begin{figure}[h]
\centering\includegraphics[width=\textwidth]{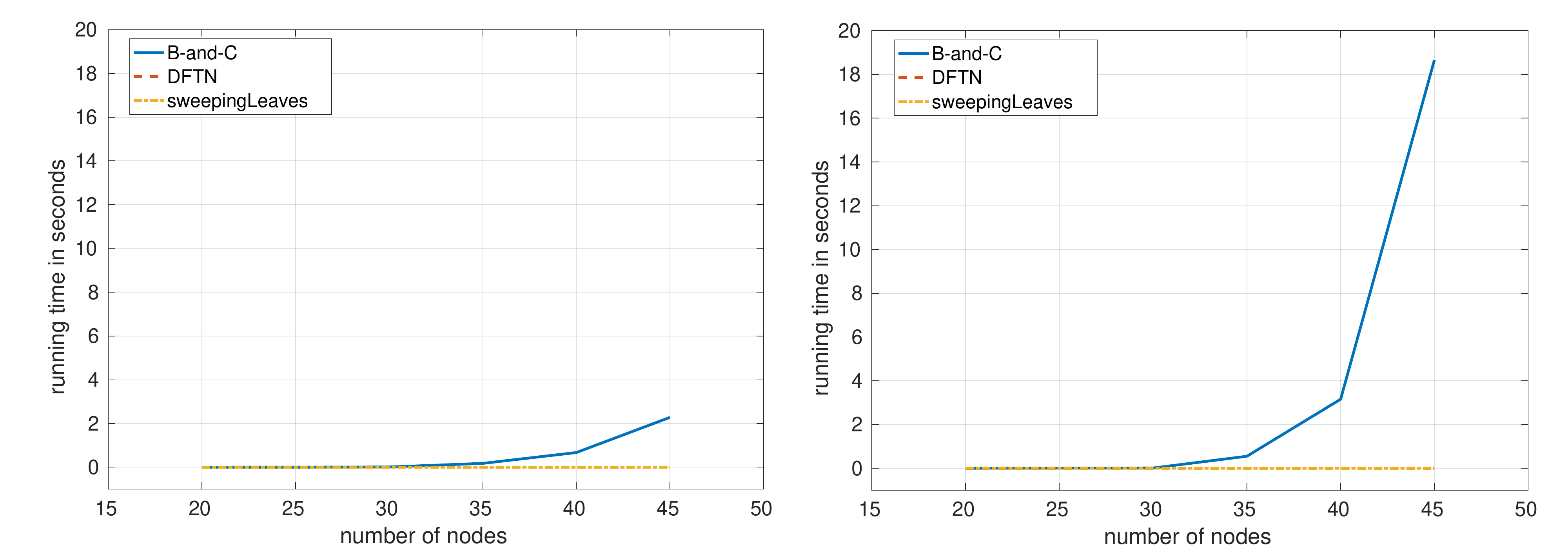}
\caption{Average running time of \texttt{deepest-first-then-nearest}, \texttt{sweepingLeaves} and \texttt{B-and-C}, \MD on the left and \MT on the right.}
\label{runtime}
\end{figure}

\begin{figure}[h]
\centering\includegraphics[width=\textwidth]{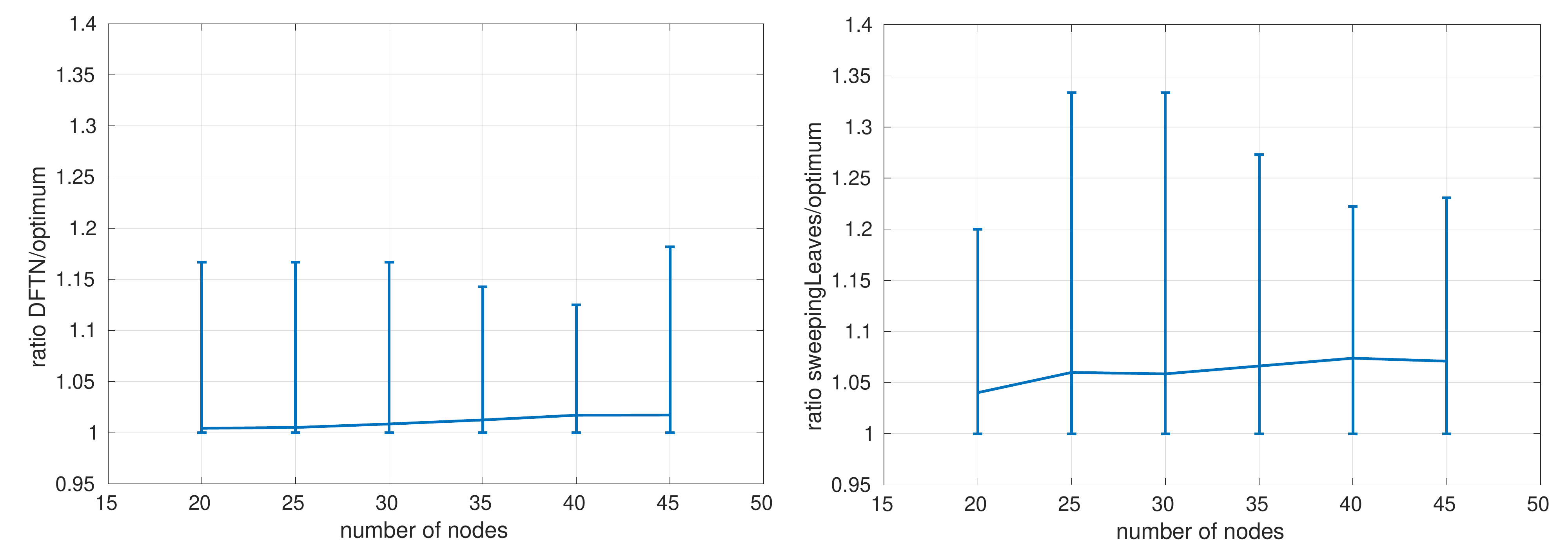}
\caption{Approximation ratio of \texttt{deepest-first-then-nearest} (left) and \texttt{sweepingLeaves} (right) respect to number of immersions.}
\label{numimmersions}
\end{figure}

Figure \ref{runtime} shows the average running time of DFTN, \texttt{deepest-first-then-} \texttt{nearest} 
and \texttt{B-and-C}. The algorithm \texttt{B-and-C} exhibits a non-polynomial running time but still is capable of solving the exact problem for $n$ up to 45. 
Finally, Figure~\ref{numimmersions} shows approximation ratios of DFTN, \texttt{deepest-first-then-nearest} 
and \texttt{B-and-C} using the number of immersions.

\section{Discussion and Open Problems}
\begin{figure}[h]
\centering\includegraphics[scale=0.9]{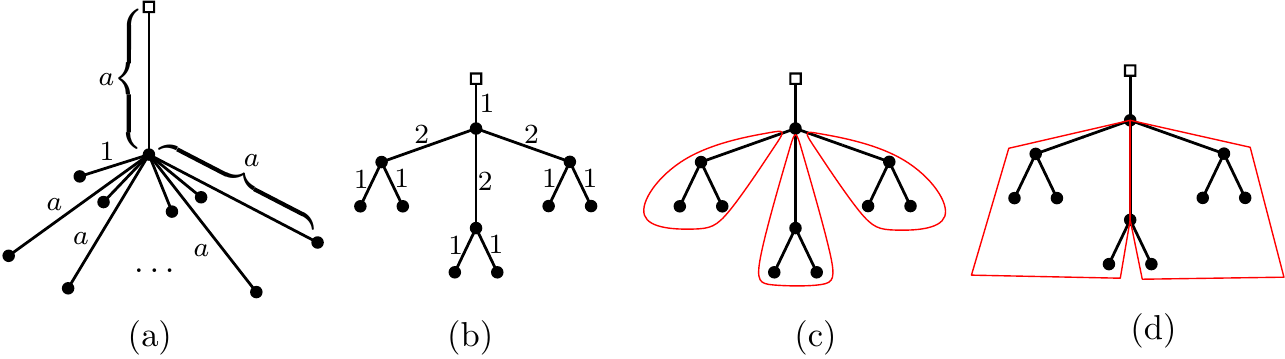}
\caption{(a) Approximation of \texttt{sweepingLeaves}. (b) An example of a tree $T$ such that \textsc{Min-Distance} and \textsc{Min-Immersions} have different solutions, (c) the solution of \textsc{Min-Distance}, and (d) the solution of \textsc{Min-Immersions}.}
\label{fig:2in1}
\end{figure}

In all our computations the approximation factor of min-Distance is always smaller than 1.2. 
The worst example that we found (not the \texttt{sweepingLeaves} program) is shown in Figure~\ref{fig:2in1}a.
The tree has a central vertex which the parent of all the leaves.
We assume that $p=2a$ in this example where $a\ge 2$ is an integer. 
The optimal solution has $a$ trees with one leaf and one tree with $a$ leaves.
The cost of each tree is $4a$.
The total cost is $a(4a+4)$.  
The solution computed by \texttt{sweepingLeaves} contains $2a$ trees, each corresponding to one leaf.
The cost of these trees is $a(2a+2)+a(4a)=a(6a+2)$. The approximation ration is 
$
\frac{6a+2}{4a+4}=\frac{3a+1}{2a+2}=\frac 32-\frac 1{a+1}.
$ 
If $a$ is large enough then the approximation ratio tends to 1.5. Based on our computational results (Figure \ref{intervals}) and preliminary investigation we conjecture that the approximation ratio is always at most 1.5.

\begin{conj}
The algorithm \texttt{sweepingLeaves} is an 1.5-approximation algorithm for \MD.
\end{conj} 

Now we elaborate on the relation between \textsc{Min-Distance} and \textsc{Min-Immer-sions}. Based on our experiments one can conjecture that there is always a common solution to both \textsc{Min-Distance} and {\textsc{Min-Immersions}.
Surprisingly, we found an example illustrated in Figure \ref{fig:2in1}b where \textsc{Min-Distance} and \textsc{Min-Immersions} have different solutions. Indeed, suppose that the input of both \textsc{Min-Distance} and \textsc{Min-Immersions} is the tree $T$ shown in Figure~\ref{fig:2in1}b and $p=16$. Then there is only one optimal solution for  \textsc{Min-Distance} that contains 3 trees of total cost 15,
see Figure~\ref{fig:2in1}c. Also, there is only one optimal solution for  \textsc{Min-Immersions} that contains 2 trees,
see Figure~\ref{fig:2in1}d. Notice that the total cost of these two trees is 16.
We believe that a different conjecture can be stated.


\begin{conj}
Any optimal solution of \textsc{Min-Immersions} has a constant approximation factor of an optimal solution of \textsc{Min-Distance}.
\end{conj} 


\bibliographystyle{plain}
\bibliography{exploringbib}

\begin{thebibliography}{1}

\bibitem{arkin2006}
E.~M. Arkin, R.~Hassin, and A.~Levin.
\newblock Approximations for minimum and min-max vehicle routing problems.
\newblock {\em Journal of Algorithms}, 59(1):1--18, 2006.

\bibitem{bektas2006}
T.~Bektas.
\newblock The multiple traveling salesman problem: an overview of formulations
  and solution procedures.
\newblock {\em Omega}, 34(3):209--219, 2006.

\bibitem{ours}
L.E. Caraballo and J.M. D\'iaz-B\'a\~nez.
\newblock Covering problems for underground robot systems.
\newblock {\em Working paper}, 2017.

\bibitem{davis2011survey}
Robert~I. Davis and Alan Burns.
\newblock A survey of hard real-time scheduling for multiprocessor systems.
\newblock {\em ACM computing surveys (CSUR)}, 43(4):35, 2011.

\bibitem{dynia2006}
M.~Dynia, M.~Korzeniowski, and C.~Schindelhauer.
\newblock Power-aware collective tree exploration.
\newblock In {\em Int. Conf. on Archit. of Computing Systems}, pages 341--351,
  2006.

\bibitem{mine}
P.~Hem and J.~Caldwell.
\newblock Block caving.
\newblock {\em Mining Technology, InfoMine}, 2012.

\bibitem{khan1994comparison}
A.A. Khan, C.L. McCreary, and M.S. Jones.
\newblock A comparison of multiprocessor scheduling heuristics.
\newblock In {\em Parallel Processing}, volume~2, pages 243--250, 1994.

\bibitem{rao1993}
N.S.V. Rao, S.~Kareti, W.~Shi, and S.S. Iyengar.
\newblock Robot navigation in unknown terrains: Introductory survey of
  non-heuristic algorithms.
\newblock Technical report, ORNL/TM-12410, Oak Ridge National Laboratory, 1993.

\end{thebibliography}
\end{document}